%% file: main.tex
\documentclass[10pt,twocolumn,letterpaper]{article}

\usepackage[pagenumbers]{iccv} % To force page numbers, e.g. for an arXiv version


\definecolor{iccvblue}{rgb}{0.21,0.49,0.74}
\usepackage[pagebackref,breaklinks,colorlinks,allcolors=iccvblue]{hyperref}
\usepackage{multirow}

\def\paperID{12173} % *** Enter the Paper ID here
\def\confName{ICCV}
\def\confYear{2025}
\def\method{FlowDreamer\xspace}

\title{\method: A RGB-D World Model with Flow-based Motion Representations for Robot Manipulation}

\author{
Jun Guo\textsuperscript{*,1,2},
Xiaojian Ma\textsuperscript{*,{\#},1},
Yikai Wang\textsuperscript{*,3},
Min Yang\textsuperscript{1,4},
Huaping Liu\textsuperscript{{\#},2},
Qing Li\textsuperscript{{\#},1} \\
\textsuperscript{1}State Key Laboratory of General Artificial Intelligence (BIGAI) \\
\textsuperscript{2}Department of Computer Science and Technology, Tsinghua University \\
\textsuperscript{3}School of Artificial Intelligence, Beijing Normal University \\
\textsuperscript{4}School of Artificial Intelligence and Automation, Huazhong University of Science and Technology \\
\textsuperscript{*}Equal Contribution \quad \textsuperscript{\#}Corresponding Author \\
\url{https://sharinka0715.github.io/FlowDreamer/}
}

\begin{document}
\maketitle

% 写作时主要参考的文章：IRASim、iVideoGPT、HMA
% 画图：Teaser figure，overview framework
\begin{abstract}

This paper investigates training better visual world models for robot manipulation, \ie, models that can predict future visual observations by conditioning on past frames and robot actions. Specifically, we consider world models that operate on RGB-D frames (RGB-D world models). As opposed to canonical approaches that handle dynamics prediction mostly implicitly and reconcile it with visual rendering in a single model, we introduce \method, which adopts 3D scene flow as explicit motion representations. \method first predicts 3D scene flow from past frame and action conditions with a U-Net, and then a diffusion model will predict the future frame utilizing the scene flow. \method is trained end-to-end despite its modularized nature. We conduct experiments on 4 different benchmarks, covering both video prediction and visual planning tasks. The results demonstrate that \method achieves better performance compared to other baseline RGB-D world models by 7\% on semantic similarity, 11\% on pixel quality, and 6\% on success rate in various robot manipulation domains. %The code and model will be released to the public.

% As opposed to canonical approaches that handle dynamics and visual appearance modeling in a single model, we introduce \method, which decouples dynamics prediction from visual generation by utilizing 3D scene flow as intermediate motion representations. 
% We hope our work encourages greater attention to spatial information and environment dynamics, driving further advancements in world modeling.

% The ability to model and predict the future is crucial for improving planning, exploration, and learning in robotics. This capability, known as world modeling, has garnered increasing attention in recent years with advancements in generative models. In this paper, we propose \method, a generative world model conditioned on robot actions, which decouples environment dynamics from generative models that primarily focus on visual fidelity. Specifically, we incorporate depth information into the world model’s observations, enabling a more comprehensive perception of 3D space. Furthermore, we leverage the 3D scene flow, a versatile motion representation, to represent the environment state transition, and introduce a dynamics prediction module that learns environment dynamics separately. To validate our approach, we conduct experiments on 4 different benchmarks, covering both video prediction and visual planning tasks. The results demonstrate that \method achieves competitive performance compared to state-of-the-art world models. We hope our work encourages greater attention to spatial information and environment dynamics, driving further advancements in world modeling.

\end{abstract}

\section{Introduction}
\label{sec:introduction}

We study developing better visual world models for robot manipulation tasks. In robotics, a visual world model~\cite{worldmodel2018} needs to perform the following steps: 1) \textbf{dynamics prediction}: predict the future motion given the current sensory observations (about robot and environment states) and robot action; and 2) \textbf{visual rendering}: render the visual observations after the motion happens. A visual world model captures the underlying world dynamics and can be used as a learnable simulator to help produce and evaluate motion plans of robot manipulators in company with model-based planning algorithms~\cite{action_unisim,action_avid,action_irasim,action_ivideogpt}. The use of visual world models alleviates the need for precise scene modeling and simulation~\cite{dai2024automated,jiang2022ditto}, making it a promising research direction. In this paper, we specifically focus on world models that operate on RGB-D frames (RGB-D world models), which are commonly adopted sensory observations in robot manipulation tasks.

\begin{figure}[!t]
\centering
\vspace{-0.2in}
\includegraphics[width=\linewidth]{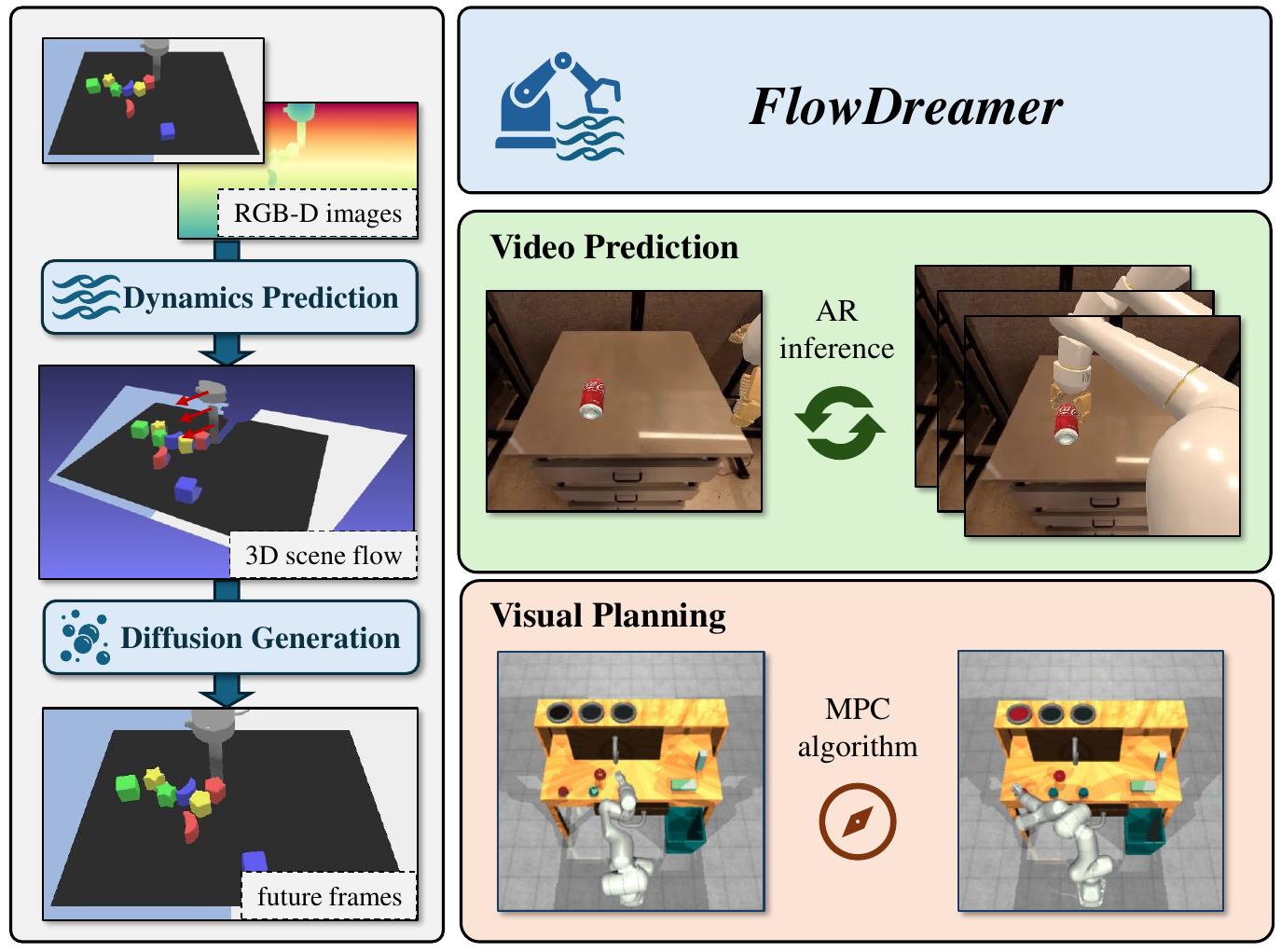}
\caption{\textbf{Proposed RGB-D world model with flow-based motion representations.} \method adopts a two-stage prediction framework, which explicitly predict scene flow as motion representations. \method achieves better results on future frame prediction and visual planning tasks in various robot manipulation domains.} 
\label{fig:teaser}
\vspace{-0.2in}
\end{figure}

Existing visual world models have undergone rapid development in recent years. Starting from early approaches that utilize recurrent neural networks (RNNs)~\cite{mbrl2018,mbrl2020,dreamerv1,dreamerv2,dreamerv3,tdmpc}, powerful diffusion-based generative models~\cite{stable_diffusion,sora,stable_diffusion_3,diffusion_ddim,diffusion_ddpm,diffusion_first} have gain popularity in recent world models~\cite{game_genie2,game_muse,text_cosmos,game_diamond}. However, regardless of the architectures of these models, they mostly reconcile the two aforementioned steps (dynamics prediction and visual rendering) in a single model. These design choices not only reduce the model’s transparency but also, as our later experiments show, impair its future prediction performance. We hypothesize that models trained solely with frame prediction loss tend to prioritize improving the fidelity of rendered visual appearances while placing less emphasis on accurate dynamics prediction. This highlights the importance of exploring methods that explicitly model dynamics prediction.

To this end, we propose \method, a RGB-D world model that explicitly models dynamics prediction to enhance the predictive capability of world models. \method adopts a two-stage framework to predict the environment dynamics and render the visual observations separately. Specifically, \method introduces explicit modeling of 3D dynamics by leveraging 3D scene flow~\cite{sceneflow_first}, which is a versatile representation that describes the motion of objects within a scene. 
In the first stage, a scene flow prediction module independently predicts the dynamic changes induced by given actions. This module is trained with a scene flow prediction loss to ensure robust supervision of the scene dynamics, thereby endowing the world model with an enhanced understanding of dynamics in 3D space. 
In the second stage, we employ a conditional diffusion model~\cite{diffusion_ddpm, diffusion_ddim} that predicts the next visual observation based on the current observation and the motion information provided by the scene flow prediction module. 
Despite its modularized nature, our model can be trained in an end-to-end fashion.
% It is noteworthy that our proposed framework is not restricted to diffusion models; alternative generative architectures, such as autoregressive Transformers~\cite{transformer, gpt}, can be readily incorporated. Consequently, our approach benefits from both additional spatial input and explicit dynamic modeling, which together improve the overall predictive capability.

% 我们在未来预测和视觉规划任务上测试了我们的方法，得到了良好的表现
We validate the effectiveness of our method on multiple benchmarks commonly used in robotic manipulation. On action-conditioned benchmarks (\eg, RT-1~\cite{rt1}, Language Table~\cite{languagetable}), our approach achieves better visual performance comparable to normal RGB-D world models, with a 7\% increase in semantic similarity and 11\% on pixel quality. Evaluations on VP$^2$ visual planning benchmark~\cite{vp2} with RoboDesk~\cite{robodesk} and Robosuite~\cite{robosuite} tasks reveal a 6\% increase in the success rate in manipulation tasks.

% 三点主要贡献
% 修改：我们是动力学解耦的RGB-D模型，同时支持端到端训练
% 我们提出scene flow的表示来建模动力学
% 实验
In summary, our main contributions are as follows:
\begin{enumerate}
    \item We propose \method, a two-stage RGB-D world model with dynamics and visual prediction module, which can be trained in an end-to-end fashion despite its modularized nature.
    \item We introduce a scene flow prediction module that adopts 3D scene flow on RGB-D space as a general motion representation to supervise dynamics prediction and enhance dynamics modeling.
    \item We perform comprehensive evaluations across several benchmarks, demonstrating the efficacy of our approach in both visual performance and visual planning tasks.
\end{enumerate}

\begin{figure*}[!t]
\centering
\vspace{-0.2in}
\includegraphics[width=\linewidth]{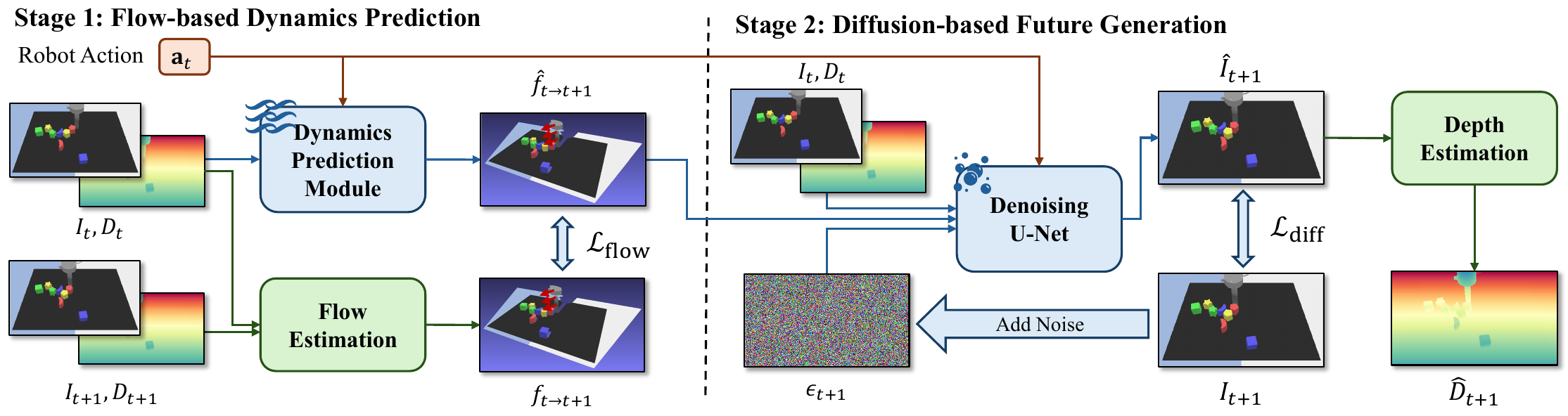}
\caption{\textbf{Overview of \method.} At stage 1, \method receives the RGB-D frame and the robot action as input to explicitly predict the scene flow as motion representations. At stage 2, \method leverages a denoising U-Net to generate high-resolution next-step future observation via diffusion.} 
\label{fig:overview}
\vspace{-0.2in}
\end{figure*}

\section{Related Work}
\label{sec:related_work}

% 生成式世界模型
\subsection{Generative World Models in Robotics}

World models~\cite{worldmodel2018}, also known as dynamic models, refer to techniques that predict the future state based on current observations and given conditions. In robotics, control strategies can leverage the generative capability of these models for planning, learning, or reducing the cost associated with interactions in the real environment. A world model typically requires the robot’s current observation in the form of latent states, sensor measurements, or visual information as input and produces future observations while also potentially predicting higher-level information such as rewards, values, or subsequent actions.

Existing work in this domain can be broadly categorized into two types, namely \textit{control-based} world models and \textit{instruction-based} world models. In \textit{control-based} world models~\cite{action_avid, action_irasim, action_ivideogpt, action_unisim, action_unsupervised, action_unsupervised2}, the conditioning inputs are robot control signals, including joint forces or velocities, end-effector positions, or movement commands. These models can serve as simulators for the environment and facilitate approaches such as model-based reinforcement learning~\cite{mbrl1991, mbrl2018, mbrl2020, mbrl2020nature, mbrl2024, dreamerv1, daydreamer, dreamerv2, dreamerv3} and model predictive control~\cite{tdmpc, tdmpc2}. Many studies~\cite{game_2015,game_deltairis,game_diamond,game_gamengen,game_genie,game_genie2,game_iris,game_muse} have also trained world models within games with players' interaction signals as input, employing neural networks as the game engine. In contrast, \textit{instruction-based} world models~\cite{text_cosmos, text_dynalang, text_enerverse, text_avdc, text_robodreamer, text_susie, text_unipi, text_vlp, action_unisim} take task-related instructions, \eg, task identifiers or natural language descriptions, as conditions to generate a sequence of future states or actions. This approach not only envisions future world states but also simulates the execution of policies, thereby allowing inverse dynamics to derive robot control parameters from observations~\cite{text_avdc, text_susie} or enabling fine-tuning into a vision-language-action (VLA) model~\cite{text_3dvla, text_gr2, text_unipi, text_vlp}. This paper proposes a \textit{control-based} world model that uses robot actions as conditions and employs scene flow as an intermediate representation to simulate and predict environmental dynamics, ultimately generating images that are interpretable by human observers.

% 运动建模
\subsection{Dynamics Modeling}

Most existing world models are trained in an end-to-end manner with supervising signals to directly predict future states without explicitly modeling the dynamical processes. Explicit modeling of the dynamics underlying environmental state changes can enhance the interpretability of the world model and improve prediction accuracy. In the field of video generation, there are several works~\cite{motion1, motion2, motion3, motion4, motion_motioni2v} that model the motion between frames to better control and enhance the performance of video generation models. Early works~\cite{t3vip, dsrnet, se3net} employed SE(3) transformations as an intermediate representation of object motion. This approach is limited to rigid bodies and requires the world model to possess object-centric awareness, which requires extra segmentations of training data. An alternative representation is \textit{flow}~\cite{im2flow2act, text_avdc}, which is referred to as optical flow~\cite{oldflow_hornschunck, oldflow_lucaskanade} in 2D images and as scene flow~\cite{sceneflow_first} in 3D space. Flow is defined as the displacement of every point in the observation space from one timestep to the next, which provides a universal and flexible means to represent various forms of motion, including non-rigid objects. In order to extract flow information from the training sequences of world models, flow estimation methods are employed to compute the displacement between adjacent frames. In the 2D domain, numerous methods, including both traditional~\cite{oldflow_hornschunck, oldflow_lucaskanade, oldflow1, oldflow2} and deep learning-based~\cite{flow1, flow2, flow3, flow4, flow5, flow6, flow7, flow_raft, track_dot} approaches, have been developed to estimate optical flow. In 3D space, several deep learning methods~\cite{sceneflow1, sceneflow2, sceneflow3, sceneflow4, sceneflow5, track_spatialtracker} have shown promise in estimating scene flow, while the task remains generally more challenging. Our approach utilizes 3D scene flow in the RGB-D space as the representation of motion, which can be obtained by integrating optical flow from consecutive frames with the corresponding depth information.

\section{\method: A RGB-D World Model}
\label{sec:method}

In this section, we illustrate the framework of our \method. Fig.~\ref{fig:overview} depicts the overall framework of our method. \method is a two-stage action-conditioned RGB-D world model, which receives the current RGB-D observation $(I_t, D_t)$ and the robot action $\mathbf{a}_t$ as the input and predicts the future RGB observations $I_{t+1}$. At stage 1, we start from an RGB-D observation image and a robot action, predicting the 3D scene flow $f_{t\rightarrow t+1}$ between the current and the next frame (Sec.~\ref{sec:dynamics_prediction}). At stage 2, we apply a diffusion denoising network condition on the RGB-D image $(I_t, D_t)$ and the predicted scene flow $\hat{f}_{t\rightarrow t+1}$ to generate the observation $I_{t+1}$ at the next timestep from a random noise $\epsilon_{t+1}$ (Sec.~\ref{sec:future_generation}).

\subsection{Background: Latent Diffusion Models}

Diffusion models~\cite{diffusion_first, diffusion_ddpm, diffusion_ddim} progressively transform data into noise through a forward diffusion process and then learn to reverse this process to generate high-quality samples. Latent diffusion models~\cite{stable_diffusion} extend this framework by performing the diffusion process in a lower-dimensional latent space rather than directly in the pixel space. By leveraging a learned encoder~\cite{vqvae} to project data into a compact latent representation, latent diffusion models achieve significant improvements in computational efficiency and scalability while still maintaining high fidelity in the generated samples. In the forward process, small amounts of Gaussian noise are incrementally added to the data, gradually destroying its structure, which can be formatted by

\begin{equation}
q(\mathbf{z}^k \mid \mathbf{z}^{k-1}) = \mathcal{N}\left(\mathbf{z}^t; \sqrt{1-\beta^k}\, \mathbf{z}^{k-1}, \beta^k \mathbf{I}\right),
\end{equation}

\noindent where $\mathbf{z}^k$ is the noised latent, $\beta^k$ is a variance schedule controlling the noise level, both at $k$-th denoising step. The reverse process can be modeled by a neural network $\epsilon_\theta$ which aims to progressively denoise the sample, recovering $\mathbf{z}^0$ from $\mathbf{z}^k$ (which is approximately Gaussian):

\begin{equation}
p_\theta(\mathbf{z}^{k-1} \mid \mathbf{z}^k) = \mathcal{N}\left(\mathbf{z}^{k-1}; \mu_\theta(\mathbf{z}^k, k), \Sigma_\theta(\mathbf{z}^k, k)\right).
\end{equation}

During training, we sample a timestep $t\in [1,K]$ and add a Gaussian noise $\epsilon^k \sim \mathcal{N}(0, \mathbf{I})$ to the clean sample $\mathbf{z}^0$ by $\mathbf{z}^k=\sqrt{\bar{\alpha}^k} \mathbf{z}^0 + \sqrt{1-\bar{\alpha}^k}\epsilon^k$, and the model is optimized to predict the added noise $\epsilon^k$, resulting in a simplified loss function:

\begin{equation}
\label{eqn:loss_diff}
\mathcal{L}_{\text{diff}} = \left\|\epsilon^k - \epsilon_\theta(\mathbf{z}^k, k)\right\|^2.
\end{equation}

In the inference phase, we can generate $\mathbf{z}^0$ by starting from a sample $\mathbf{z}^K$ drawn from a standard Gaussian distribution, and the model iteratively computes $\mathbf{z}^{k-1}$ from $\mathbf{z}^k$:

\begin{equation}
\mathbf{z}^{k-1} = \frac{1}{\sqrt{\alpha^k}}\left(\mathbf{z}^k - \frac{1-\alpha^k}{\sqrt{1-\bar{\alpha}^k}}\, \epsilon_\theta(\mathbf{z}^k, t)\right) + \sigma^k \varepsilon
\end{equation}

\noindent where $\varepsilon \sim \mathcal{N}(0, \mathbf{I})$ and $\sigma^k$ is an appropriately chosen noise scale. In practice, we can accelerate the generation process by setting $\sigma^k=0$ and use a sub-sequence of $[1,\dots, K]$ to reduce the forward steps~\cite{diffusion_ddim}.

\subsection{Dynamics Prediction}
\label{sec:dynamics_prediction}

Different from end-to-end world models, \method applies a dynamics prediction module to explicitly predict the transition between consecutive frames. In robot manipulation tasks, we hope to find an intermediate representation as auxiliary information to depict the dynamics of the robot and objects. Innovated by \cite{dsrnet, text_avdc, im2flow2act}, we choose 3D scene flow as the intermediate representation, which is general and versatile and can be collected by various flow estimation networks. In point clouds, scene flow is generally defined by the displacement of point coordinates. We can easily project an RGB-D image into point cloud space with camera intrinsics, where 2D pixels in the image with shape $(H, W)$ are converted to $H\times W$ points in 3D space. We denote $K$ as the camera intrinsic matrix, $(u, v)$ as the horizontal and vertical index of pixels, and $D$ as the depth map. Then, the pixel coordinates in camera coordinate space $(x,y,z)$ can be calculated by:

\begin{equation}
    \begin{bmatrix} x \\ y \\ z \end{bmatrix} 
    = D(u, v) \cdot K^{-1}  
    \begin{bmatrix} u \\ v \\ 1 \end{bmatrix}.
\end{equation}

Assume the point in 3D space has a coordinate $(x,y,z)$ at timestep $t$ and $(x',y',z')$ at timestep $t+1$, the 3D scene flow is defined as follows:

\begin{equation}
    f_{t\rightarrow t+1} = (x'-x, y'-y,z'-z).
\end{equation}

As pixels at the same index between consecutive frames do not always represent the same 3D point, the vital challenge to get the scene flow is to find the correspondence between frames. For simulation data, we directly obtain the scene flow from the simulator backend according to the rigid transformations of every object. For real-world data, we can apply an off-the-shelf 3D scene flow estimator named RAFT-3D~\cite{flow_raft3d} to estimate the scene flow. If the real-world data has no depth information, we can estimate the depth by a pretrained video depth estimator~\cite{video_depth_anything}. The detailed scene flow obtainment pipeline are introduced in Appendix~\ref{sec:data_collection}.

We apply a dynamics prediction module to predict the 3D scene flow from time $t$ to $t+1$, given the RGB-D observation $(I_t, D_t)$ and the robot action $\mathbf{a}_t$. The backbone of our dynamics prediction model is a conditional U-Net~\cite{unet}, and the robot action is integrated with the feature map via cross-attention~\cite{transformer}. The loss function $\mathcal{L}_\text{flow}$ is defined as the mean square error (MSE) between the module output $\hat{f}_{t\rightarrow t+1}$ and the estimated scene flow $f_{t\rightarrow t+1}$:

\begin{equation}
    \label{eqn:loss_flow}
    \mathcal{L}_\text{flow} = \text{MSELoss}(\hat{f}_{t\rightarrow t+1} - f_{t\rightarrow t+1}).
\end{equation}

\subsection{Future Generation}
\label{sec:future_generation}

After getting the predicted scene flow, we can further predict future observations through a generative model. We choose Latent Diffusion Models as our future generation model, as it has shown outstanding capability in image generation~\cite{stable_diffusion, stable_diffusion_3}. 
In \method, we fine-tune the pre-trained Stable Diffusion~\cite{stable_diffusion} to build our future generation module, which we denote as $\epsilon_\theta$. The input of the generation module contains the current RGB-D observation $(I_t, D_t)$, the robot action $\mathbf{a}_t$, and the predicted 3D scene flow $\hat{f}_{t\rightarrow t+1}$, and the output of the module is the next RGB observation $I_{t+1}$. In order to generate high-resolution observations, following previous works~\cite{action_irasim, action_avid}, we use the pre-trained variational autoencoder in Stable Diffusion~\cite{stable_diffusion} to compress the RGB observation $I_t$ into latent space $\mathbf{z}_{t}$. The module output $\mathbf{z}_{t+1}$ is also in latent space and can be decoded to image space by the pretrained decoder. Depth map $D_t$ and the scene flow $\hat{f}_{t\rightarrow t+1}$ are firstly downsampled to the same shape of $\mathbf{z}_{t}$ by several convolutional layers, and then channel-wise concatenated with $\mathbf{z}_{t}$ as the input of the diffusion model:

\begin{align}
    \mathbf{z}_t &= \text{VAE}(I_t), \\
    \mathbf{c}_t &= \text{downsample}(D_t, \hat{f}_{t\rightarrow t+1}), \\
    \hat{\epsilon}^k &= \epsilon_\theta(\text{concat}(\mathbf{z}^k_{t+1}, \mathbf{z}_t, \mathbf{c}_t), \mathbf{a}_t, k),
\end{align}

\noindent where $k$ is the diffusion step, and $z^k$ is the latent to be denoised. We do not directly predict the depth map $D_{t+1}$, as the decoder of the diffusion model is not specially designed for metric depths. To autoregressively imagine the future, we leverage a pre-trained depth estimation model~\cite{depth_anything_v2} with a simple DPT~\cite{dpt} head to predict the metric depth $D_{t+1}$ from $I_{t+1}$.

In summary, our \method trains the dynamics prediction module and the denoising U-Net jointly, and the overall loss function is:

\begin{equation}
\mathcal{L}_{\text{total}} = \mathcal{L}_{\text{diff}} + \alpha \mathcal{L}_{\text{flow}},
\end{equation}

\noindent where $\mathcal{L}_{\text{diff}}$ and $\mathcal{L}_{\text{flow}}$ are defined in Eqn.~\ref{eqn:loss_diff} and Eqn.~\ref{eqn:loss_flow}, and $\alpha$ is a coefficient to control the weight of the two objectives.

\begin{table*}[!t]
\centering
\caption{\textbf{Video prediction results on the SimplerEnv RT-1 benchmark.} We categorize the metrics into three groups: semantic similarity, pixel similarity, and media quality. \textbf{Bold} numbers indicate the best results, and \underline{underlined} numbers indicate the second best results.}
\label{tab:video_prediction_rt}

\setlength{\tabcolsep}{11pt}
\begin{tabular}{@{}lcc|ccc|cc@{}}
\toprule
\multirow{2}{*}{\textbf{Method}} & \multicolumn{2}{c|}{\textbf{Semantic Similarity}} & \multicolumn{3}{c|}{\textbf{Pixel Similarity}}                                      & \multicolumn{2}{c}{\textbf{Media Quality}}     \\
                                 & DINOv2 L2↓          & CLIP score↑           & PSNR↑            & SSIM↑           & LPIPS↓          & FID↓             & FVD↓              \\ \midrule
Vanilla                          & 12.5936             & 0.8999             & 20.5925          & 0.7831          & 0.1304          & 71.8069          & 365.0697          \\
MinkNet                          & 12.1569             & 0.9038              & 20.5804          & 0.7942          & 0.1252          & 57.9706          & 325.3416          \\
SepTrain                         & \underline{11.4512} & \underline{0.9131} & \underline{21.2289} & \underline{0.8135}          & \underline{0.1097}          & \underline{45.3967}          & \textbf{245.9106} \\
\method                          & \textbf{10.9922}    & \textbf{0.9189}    & \textbf{21.7562} & \textbf{0.8196} & \textbf{0.0993} & \textbf{43.5807} & \underline{268.3853}          \\ \bottomrule
\end{tabular}
\end{table*}

\begin{table*}[!t]
\centering
\caption{\textbf{Video prediction results on the Language Table benchmark.} We categorize the metrics into three groups: semantic similarity, pixel similarity, and media quality. \textbf{Bold} numbers indicate the best results, and \underline{underlined} numbers indicate the second best results.}
\label{tab:video_prediction_lt}

\setlength{\tabcolsep}{12.5pt}
\begin{tabular}{@{}lcc|ccc|cc@{}}
\toprule
\multirow{2}{*}{\textbf{Method}} & \multicolumn{2}{c|}{\textbf{Semantic Similarity}} & \multicolumn{3}{c|}{\textbf{Pixel Similarity}}                                      & \multicolumn{2}{c}{\textbf{Media Quality}}     \\
                                 & DINOv2 L2↓              & CLIP score↑                & PSNR↑            & SSIM↑           & LPIPS↓          & FID↓             & FVD↓                   \\ \midrule
Vanilla                          & 10.7294                 & 0.9473                  & 25.6762          & 0.9273          & 0.0627          & 26.6415          & 110.4629               \\
MinkNet                          & \underline{10.0136}           & \underline{0.9614}            & 25.1973          & 0.9228          & 0.0642          & 33.5479          & 88.3917                \\
SepTrain                         & 10.2514                 & 0.9507                  & \underline{25.9785}    & \underline{0.9278}    & \underline{0.0571}    & \underline{21.2822}    & \underline{87.3121}          \\
Ours                             & \textbf{9.3899}         & \textbf{0.9688}         & \textbf{26.8907} & \textbf{0.9401} & \textbf{0.0476} & \textbf{20.0259} & \textbf{66.9215}       \\ \bottomrule
\end{tabular}
\end{table*}

\section{Experiments}
\label{sec:experiments}

In this section, we conduct extensive experiments in four different benchmarks to verify the performance of \method. We aim to answer three questions: 
\begin{enumerate}
    \item Is \method effective on video prediction compared with other RGB-D world models? (Sec.~\ref{sec:video_prediction})
    \item Can \method facilitate model predictive control for robot manipulation tasks? (Sec.~\ref{sec:visual_planning})
    \item How does the predicted scene flow in \method contribute to the future prediction? (Sec.~\ref{sec:analysis})
\end{enumerate}

\subsection{Video Prediction}
\label{sec:video_prediction}

To evaluate the future generation performance of \method, we conduct the video prediction experiments, which require the world model to generate a full trajectory given the first frame of the video and the action trajectory. By comparing the similarity between predicted frames and ground truth frames, we can verify the future prediction capability of world models.

\noindent\textbf{Datasets.} We conduct video prediction experiments on RT-1~\cite{rt1} and Language Table~\cite{languagetable} benchmarks to evaluate the methods. As the real-world data do not contain the depth information, we collect training and inference trajectories from the simulator. For RT-1 environment, we use SimplerEnv~\cite{simplerenv} as the simulator. For the Language Table environment, we use the official simulation environment. Collected trajectories are split into training, validation, and test sets without overlap. Experiments on real-world data are reported in Appendix~\ref{sec:extended_experiments}.

\begin{figure*}[!t]
\centering
\includegraphics[width=1.0\linewidth]{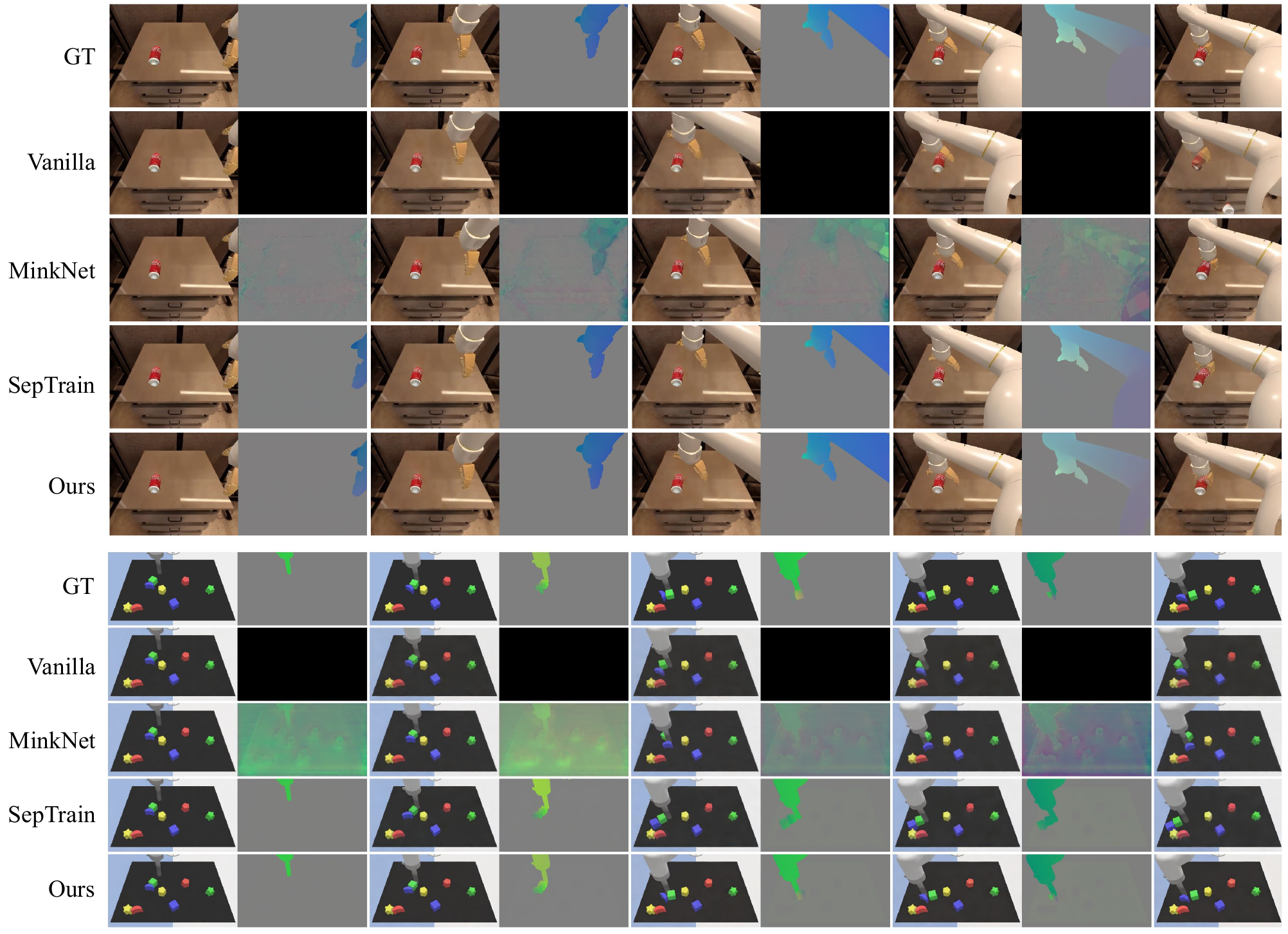}
\caption{\textbf{Qualitative results on the SimplerEnv RT-1 and Language Table benchmark.} We show the predicted frames and the scene flows except for \textit{Vanilla}, where only RGB frames are being predicted. The R, G, and B channel values in the flow visualization represent the components of the 3D scene flow along the $x$, $y$, and $z$ directions, respectively, normalized by the maximum value of the scene flow.} 
\label{fig:qualitative_rt1}
\vspace{-0.2in}
\end{figure*}

\noindent\textbf{Baselines.} To evaluate the performance of \method compared to other RGB-D world models, we design three different baselines: \textit{Vanilla}, \textit{MinkNet}, and \textit{SepTrain}. See detailed implementation in Appendix~\ref{sec:imp_details}.
\begin{itemize}
    \item \textit{Vanilla} is a single-stage RGB-D diffusion model, which receives current RGB-D observations as input and predicts RGB images at the next timestep. We compare it with our \method to measure the contribution of the dynamics prediction module. 
    \item \textit{MinkNet} is a two-stage world model that replaces the backbone of the dynamics prediction module from U-Net to MinkowskiNet~\cite{minknet}, a 4D sparse convolutional network for point clouds. We compare it with our \method to demonstrate the effectiveness of the RGB-D representation. 
    \item \textit{SepTrain} is a two-stage world model, which trains the dynamics prediction module and the Denoising U-Net separately. We compare it with our \method to evaluate the performance of end-to-end training.
\end{itemize}

\noindent\textbf{Metrics.} To evaluate the video prediction performance, we use PSNR~\cite{psnr}, SSIM~\cite{ssim}, LPIPS~\cite{lpips}, FID~\cite{fid}, and FVD~\cite{fvd} as the assessment metrics, and additionally calculate the latent L2 distance extracted by DINOv2 (denoted as DINOv2 L2) and the latent cosine similarity extract by CLIP (denoted as CLIP score) to estimate the semantic similarity. PSNR measures the distance between the predicted video and the ground-truth video in the pixel space, and SSIM evaluates the structural similarity between frames. LPIPS, FID, and FVD are model-based evaluation metrics that compare frames in latent feature space. LPIPS measures the distance in different feature spaces, while FID and FVD measure the distribution disparity between generated and ground-truth images or videos. DINOv2 and CLIP are large-scale pretrained models via unsupervised learning, which could robustly extract the semantic feature from the images. We assume that DINOv2 and CLIP feature distance could reflect the environment state information more than image quality metrics.

\noindent\textbf{Results.} Table~\ref{tab:video_prediction_rt} and \ref{tab:video_prediction_lt} shows the quantitative results on RT-1 and Language Table benchmarks. We can observe that our \method achieves the best performance on most of the metrics, including semantic similarity, pixel similarity, and media quality. The results demonstrate the effectiveness of our \method. The \textit{SepTrain} model achieves the second-best performance at most of the metrics, and the performance is very similar to our \method. This indicates that end-to-end training is generally a better approach, while the contribution is less than that of other components. We notice that \textit{Vanilla} and \textit{MinkNet} have similar performances, though \textit{MinkNet} has a similar two-stage framework and a dynamics prediction module. From the qualitative results (Fig.~\ref{fig:qualitative_rt1}), we can see the scene flow predicted by \textit{MinkNet} has low quality. The color of the predicted flows is similar to the ground truth flows, which shows that the direction of flows is generally correct. However, the flow predicted by \textit{MinkNet} cannot distinguish the moving parts and the stationary parts, thus not providing efficient guidance to the downstream diffusion model. Moreover, we observe from Fig.~\ref{fig:qualitative_rt1} that all predictions have a similar visual quality, but our \method outperforms other models in environment states. This lies in the additional dynamics prediction module, which can explicitly predict the environment transition, thus resulting in more accurate future predictions.

\begin{figure*}[!t]
\centering
\includegraphics[width=\linewidth]{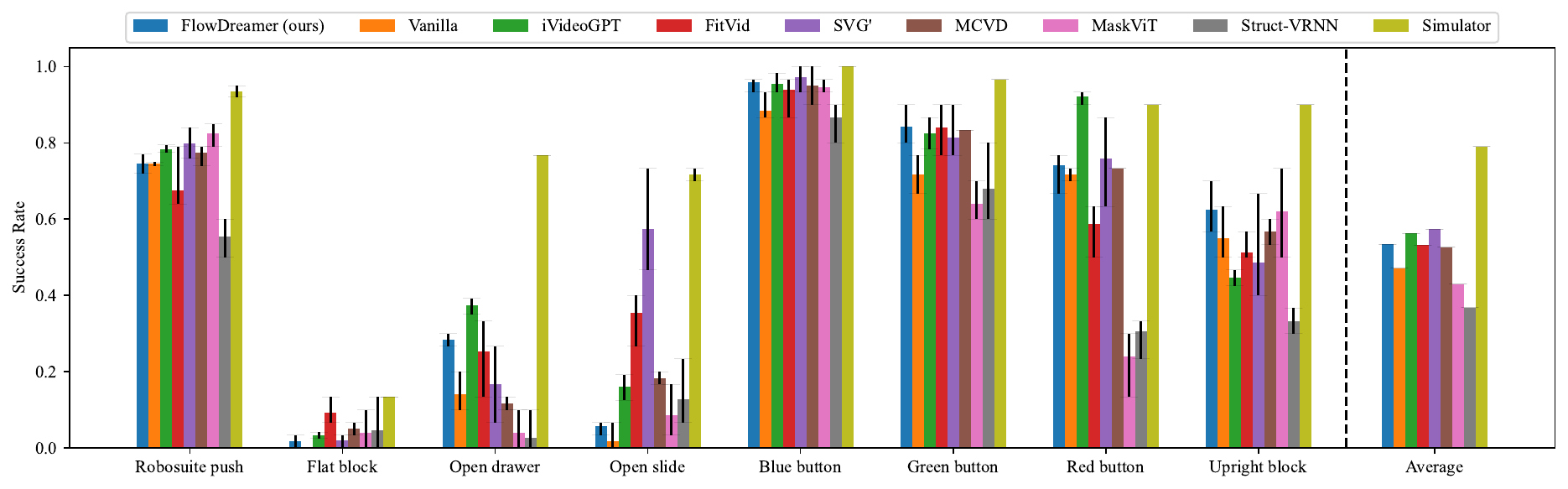}
\caption{\textbf{Visual planning results on the VP$^2$ benchmark.} We report the mean and the min/max performance of different methods over multiple runs with different random seeds. On the right, ``Average'' means the average success rate over all reported tasks.} 
\label{fig:visual_planning}
\end{figure*}

\begin{figure*}[!t]
\centering
\includegraphics[width=1.0\linewidth]{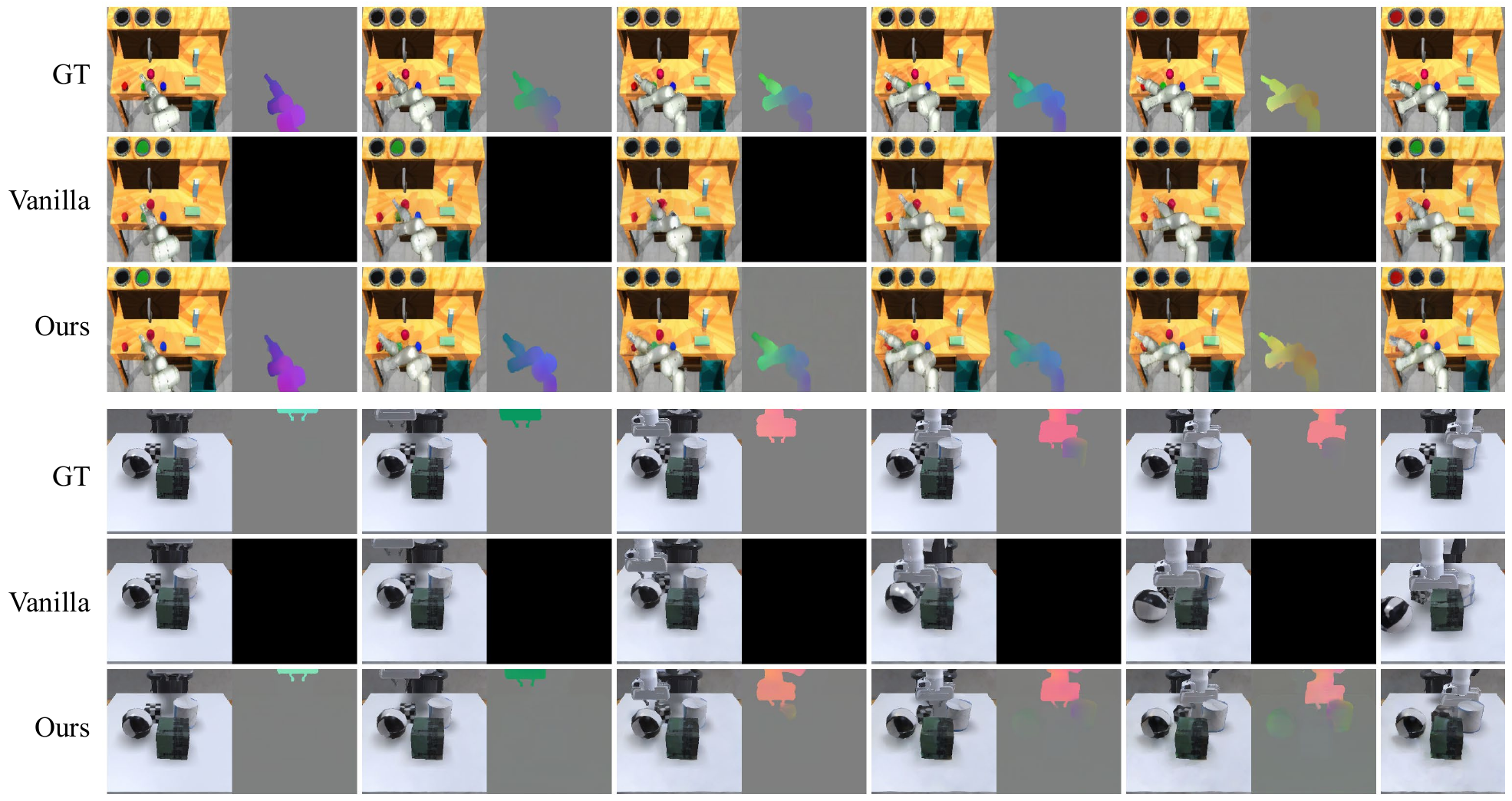}
\caption{\textbf{Qualitative results on the Robodesk and Robosuite dataset.} The trajectory comes from the validation set, which is split from the original training trajectories and is not used for training. For our method, we show the predicted RGB images and scene flows.} 
\label{fig:qualitative_visual_planning}
\vspace{-0.2in}
\end{figure*}

\subsection{Visual Planning}
\label{sec:visual_planning}

We further evaluate our \method on visual planning tasks to show how \method makes a difference in robot manipulation tasks. In visual planning tasks, the policy interacts with environments to minimize the difference between the observation and the goal image. For world models without any action output, model predictive control (MPC)~\cite{action_unsupervised2, mbrl2018} methods can be applied to evaluate the performance.

\noindent\textbf{Datasets.} We choose VP$^2$~\cite{vp2} as our visual planning benchmark. VP$^2$ is a control-centric benchmark that evaluates video prediction models by visual MPC methods. The environment contains four Robosuite~\cite{robosuite} and seven RoboDesk~\cite{robodesk} tasks. We run our experiments with 4 seeds on RoboDesk tasks and 3 seeds with Robosuite tasks, which keeps same with other VP$^2$ experiments.

\noindent\textbf{Baselines.} We compare the visual planning performance with a single-stage RGB-D diffusion world model (denoted as \textit{Vanilla}). Moreover, following iVideoGPT~\cite{action_ivideogpt}, we choose all video generation models provided by VP$^2$ paper as our baselines, including FitVid~\cite{fitvid}, SVG'~\cite{svg}, MCVD~\cite{mcvd}, Struct-VRNN~\cite{structvrnn}, and MaskViT~\cite{maskvit}. We also compare our performance with iVideoGPT itself.

\noindent\textbf{Metrics.} Following iVideoGPT~\cite{action_ivideogpt}, we report the minimum, maximum, and average success rate of our method between different random seeds. The reported baseline results are provided by previous works~\cite{action_ivideogpt, vp2}, and we additionally report our performance in the same setting. For Robosuite push tasks, a cost below 0.05 is considered a success.

\begin{figure*}[!ht]
    \centering
    \includegraphics[width=1.0\linewidth]{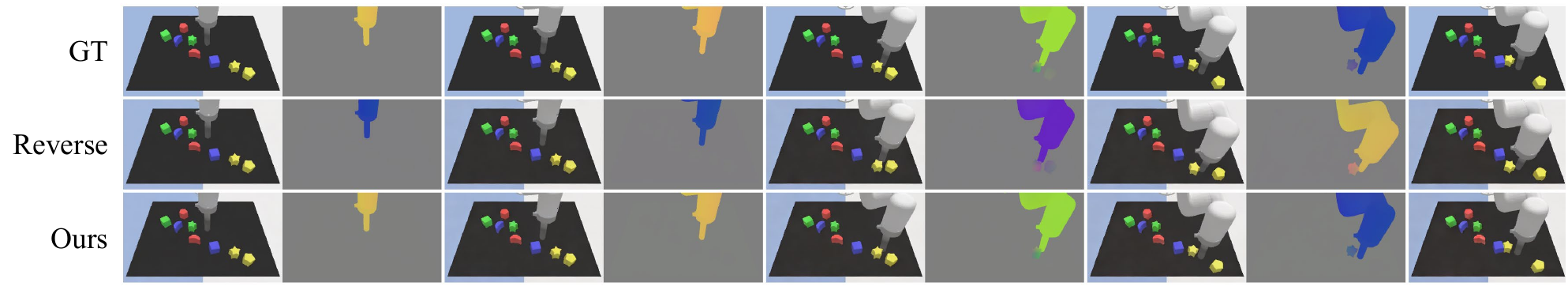}
    \caption{\textbf{The qualitative results when flows are reversed.} With reversed (therefore incorrect) scene flow, the diffusion model in \method can only utilize action condition, leading to worse performances on future frame prediction.}
    \label{fig:reverse}
    \vskip -0.1in
\end{figure*}

\begin{figure}[!h]
    \centering
    \begin{subfigure}[t]{0.48\linewidth}
        \centering
        \includegraphics[width=\linewidth]{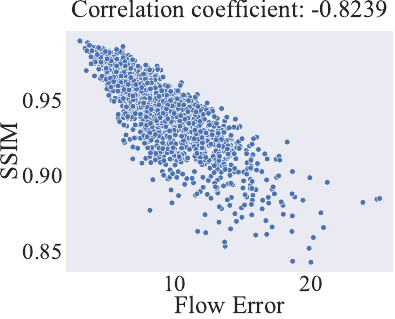}
        \caption{SSIM}
    \end{subfigure}
    \begin{subfigure}[t]{0.48\linewidth}
        \centering
        \includegraphics[width=\linewidth]{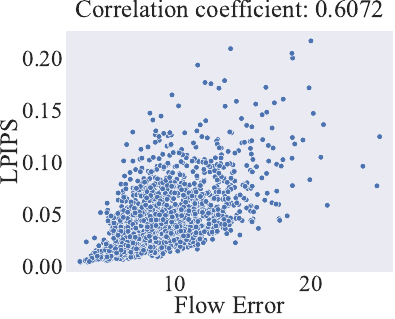}
        \caption{LPIPS}
    \end{subfigure}
    \begin{subfigure}[t]{0.48\linewidth}
        \centering
        \includegraphics[width=\linewidth]{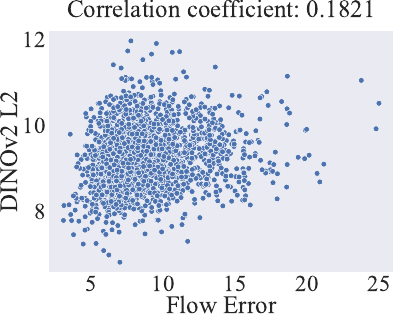}
        \caption{DINOv2 L2}
    \end{subfigure}
    \begin{subfigure}[t]{0.48\linewidth}
        \centering
        \includegraphics[width=\linewidth]{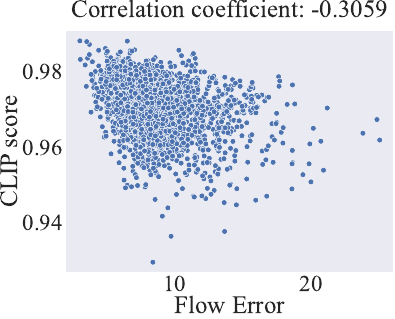}
        \caption{CLIP score}
    \end{subfigure}
    \caption{\textbf{The correlation between the flow prediction error and image assessment metrics.} We show the scatter plots of SSIM (higher is better), LPIPS (lower is better), DINOv2 L2 (lower is better), and CLIP score (higher is better) vs. flow error and report the correlation coefficients.}
    \label{fig:correlation}
    % \vspace{-0.1in}
\end{figure}

\begin{table}[!h]
    \centering
    \setlength{\tabcolsep}{3pt}
    \begin{tabular}{@{}lccccc@{}}
    \toprule
    $r$         & PSNR    & SSIM    & LPIPS  & DINOv2 & CLIP    \\ \midrule
    Flow Error  & -0.6704 & -0.8239 & 0.6072 & 0.1821 & -0.3058 \\ \bottomrule
    \end{tabular}
    \caption{\textbf{The correlation coefficient $r$ between the flow prediction error and image assessment metrics.} $r>0$ indicates a positive correlation and $r<0$ indicates a negative correlation.}
    \label{tab:correlation}
    \vspace{-0.2in}
\end{table}

\noindent\textbf{Results.} Fig.~\ref{fig:visual_planning} shows the visual MPC results on the VP$^2$ benchmark. From the results, we found that our \method always performs better than the \textit{Vanilla} model, which implies that our proposed framework works well on visual planning tasks. For other video prediction models, \method achieves a comparable performance on the average metric, while all video prediction baselines have a context of two frames. Our \method performs best on some tasks, \eg, Green button and Upright block, while it fails on other tasks, \eg, Open slide. We hypothesize that the failure lies in that the visual reward cannot always point to the correct trajectory, which is also revealed by \cite{action_ivideogpt}. Fig.~\ref{fig:qualitative_visual_planning} shows the qualitative results on VP$^2$ benchmark tasks. Our model can predict the flow and the goal accurately on RoboDesk tasks, while it is hard to predict precise flow on Robosuite push tasks. We assume the reason lies in the lack of historical context. In Robosuite tasks, objects would have velocity after being pushed by the robot arm, which cannot be reflected by a single RGB-D image. Those objects without attachment to the robot arm have scene flows, which could be misleading in prediction.

\subsection{Additional Analysis on Flow Prediction}
\label{sec:analysis}

In this section, we conduct further analysis to figure out the effect of the predicted flow. We first reverse the direction of input flows at stage 2 while the robot action remains unchanged. Fig.~\ref{fig:reverse} visualizes the resulting prediction. We can observe that the robot did not really take contrary actions due to the action input at stage 2, while its performance becomes worse and cannot lead to the goal state. This result shows that the scene flow predicted at stage 1 provides auxiliary information to better generate the future.

Then, we calculate the correlation between the flow prediction error and other image assessment metrics. Table~\ref{tab:correlation} shows the correlation coefficient $r$ between flow prediction error and other metrics, and Fig.~\ref{fig:correlation} shows the scatter plot for SSIM, LPIPS, DINOv2 L2, and CLIP score. We notice that the flow error has a high correlation with PSNR and SSIM and a reasonable correlation with LPIPS, which demonstrates the effectiveness of the predicted flow in video prediction. For DINOv2 L2 and CLIP scores, the correlations are weak, where we infer that the semantic metrics extracted by DINOv2 and CLIP do not tell the relatively minor prediction error from the ground truth well. 

\section{Conclusion}
\label{sec:conclusion}

We introduce \method, an action-conditioned RGB-D world model with flow-based motion representations. \method leverages 3D scene flow as a versatile motion representation and applies a separate dynamics prediction module to learn environment dynamics from scene flow. Despite its modularized nature, \method can be jointly trained in an end-to-end manner. Experiments on 4 different benchmarks, including video prediction and visual planning, demonstrate the superiority of our \method compared to other RGB-D world models. Limitations and future directions can be found in the Appendix.

{
    \small
    \bibliographystyle{ieeenat_fullname}
    \bibliography{reference_header,main}
}

\clearpage
\appendix
\section{Implementation Details}  
\label{sec:imp_details}

\noindent\textbf{Details of the dynamics prediction module.} Our \method utilizes a conditional 2D U-Net with 4 downsample and upsample layers as the dynamics prediction module. The action condition is integrated with image features via cross-attention, which is similar to Stable Diffusion. We also design a conditional MinkowskiNet~\cite{minknet} as a point-cloud-based baseline. The structure of the conditional MinkowskiNet is modified from MinkUNet34B in the official repository, where we replace the batch normalization layers in the basic blocks with Feature-wise Linear Modulation (FiLM)~\cite{film} layers to introduce action condition.

\noindent\textbf{Training and inference details.} We conduct all our experiments on an NVIDIA A800 cluster. Each model is trained on 8 GPUs in parallel, with a batch size of 16 per GPU. The denoising U-Net and the image encoder/decoder are loaded from Stable Diffusion 2.1~\cite{stable_diffusion}. We only finetune the parameters of the denoising network yet freeze the weight of the encoder and decoder. All models are trained for 60k steps with a constant learning rate of 1e-4. We use AdamW optimizer for training, and we use a mixed precision with FP16 and FP32 supported by Pytorch-Lightning. For diffusion models, we utilize a PNDM scheduler with 20 sampling steps during inference. We notice that increasing sampling steps more than 20 cannot improve the accuracy of future prediction yet is more time-consuming. The output resolution is relevant to the dataset, which will be reported in the next section.

\section{Data Collection}
\label{sec:data_collection}

In the main paper, we conduct experiments on 4 different simulation environments. To train world models on these environments, we need to collect feasible trajectories and calculate the ground truth 3D scene flow. Also, we conduct experiments on RT-1 real dataset, where we estimate the depth and the scene flow through depth estimators and flow estimators. Table~\ref{tab:dataset_statistics} shows the dataset statistics in detail. This section will introduce the data collection pipeline.

\begin{table}[!h]
\centering
\begin{tabular}{@{}lccc@{}}
\toprule
Dataset        & Resolution & Episode & Sample    \\ \midrule
RT-1 SimplerEnv & $320\times 256$   & 25,000  & 758,015   \\
Language Table  & $512\times 288$   & 36,000  & 1,447,805 \\
RoboDesk        & $320\times 320$   & 35,000  & 1,190,000 \\
Robosuite       & $256\times 256$   & 50,000  & 1,749,825 \\
RT-1 Real       & $320\times 256$   & 86,403  & 3,638,016 \\ \bottomrule
\end{tabular}
\caption{\textbf{Dataset statistics.} An ``episode'' refers to a complete trajectory where the robot completes a task. A ``sample'' refers to a frame pair at timestep $t$ and $t+1$ with a robot action $\textbf{a}_t$.}
\label{tab:dataset_statistics}
\end{table}

\begin{table*}[!t]
\centering
\caption{\textbf{Video prediction results on RT-1 real-world dataset.} We categorize the metrics into three groups: semantic similarity, pixel similarity, and media quality. \textbf{Bold} numbers indicate the best results.}
\label{tab:video_prediction_real}

\setlength{\tabcolsep}{11pt}
\begin{tabular}{@{}lcc|ccc|cc@{}}
\toprule
\multirow{2}{*}{\textbf{Method}} & \multicolumn{2}{c|}{\textbf{Semantic Similarity}} & \multicolumn{3}{c|}{\textbf{Pixel Similarity}}                                      & \multicolumn{2}{c}{\textbf{Media Quality}}     \\
                                 & DINOv2 L2↓          & CLIP score↑           & PSNR↑            & SSIM↑           & LPIPS↓          & FID↓             & FVD↓              \\ \midrule
Vanilla & 15.6659         & 0.8618          & 17.9724          & 0.5401          & 0.1882          & 13.1636          & 195.3965          \\
Ours    & \textbf{15.5710} & \textbf{0.8687} & \textbf{18.6574} & \textbf{0.5532} & \textbf{0.1764} & \textbf{10.4547} & \textbf{191.3314}         \\ \bottomrule
\end{tabular}
\end{table*}

\noindent\textbf{Trajectory collection.} For the RT-1 SimplerEnv environment, we choose 5 tasks implemented by SimplerEnv~\cite{simplerenv}: \textit{pick\_coke\_can, pick\_object, move\_near, open\_drawer, close\_drawer}, and generate 5k trajectories for each task. The action trajectory is generated by CogACT~\cite{cogact}, a foundational vision-language-action model. We only sample successful trajectories and discard failed trajectories. For the Language Table environment, we generate trajectories by the RRT* oracle policy provided in the official repositories. The task instruction format is ``\textit{push \{block A\} to \{block B\}}''. For RoboDesk and Robosuite tasks, we use the trajectories provided by VP$^2$ benchmark~\cite{vp2}. We load the provided initial states and actions into the simulator to collect RGB-D and flow information. Note that the observation of RoboDesk is a resized and cropped image from the original observation, which would affect the flow calculation. Our world model trains and predicts based on the original observation and crops the predicted observation to calculate the visual planning reward. The RT-1 real-world dataset is provided by the original paper~\cite{rt1}, and we choose the preprocessed version by IRASim~\cite{action_irasim}.

\noindent\textbf{Scene flow obtainment.} For all simulation environments, we can extract the ground truth scene flow from the simulator. Specifically, all objects in the simulator are rigid bodies or articulated objects, whose \textit{pose} at each timestep $t$ could be represented by a $4\times 4$ homogeneous transformation matrix $T_t$. Therefore, we can calculate the position of a point at timestep $t+1$, with the position at timestep $t$ and the corresponding pose matrices:

\begin{equation}
    \begin{pmatrix} \mathbf{x}_{t+1} \\ 1 \end{pmatrix}
     = T_{t+1}T^{-1}_t \begin{pmatrix} \mathbf{x}_t \\ 1 \end{pmatrix},
\end{equation}

\noindent where $\mathbf{x}_{t}$ and $\mathbf{x}_{t+1}$ are 3D point coordinates at timestep $t$ and $t+1$. Then the scene flow can be calculated by $f_{t\rightarrow t+1}=\mathbf{x}_{t+1}-\mathbf{x}_{t}$.

\begin{figure}[!h]
\centering
\includegraphics[width=1.0\linewidth]{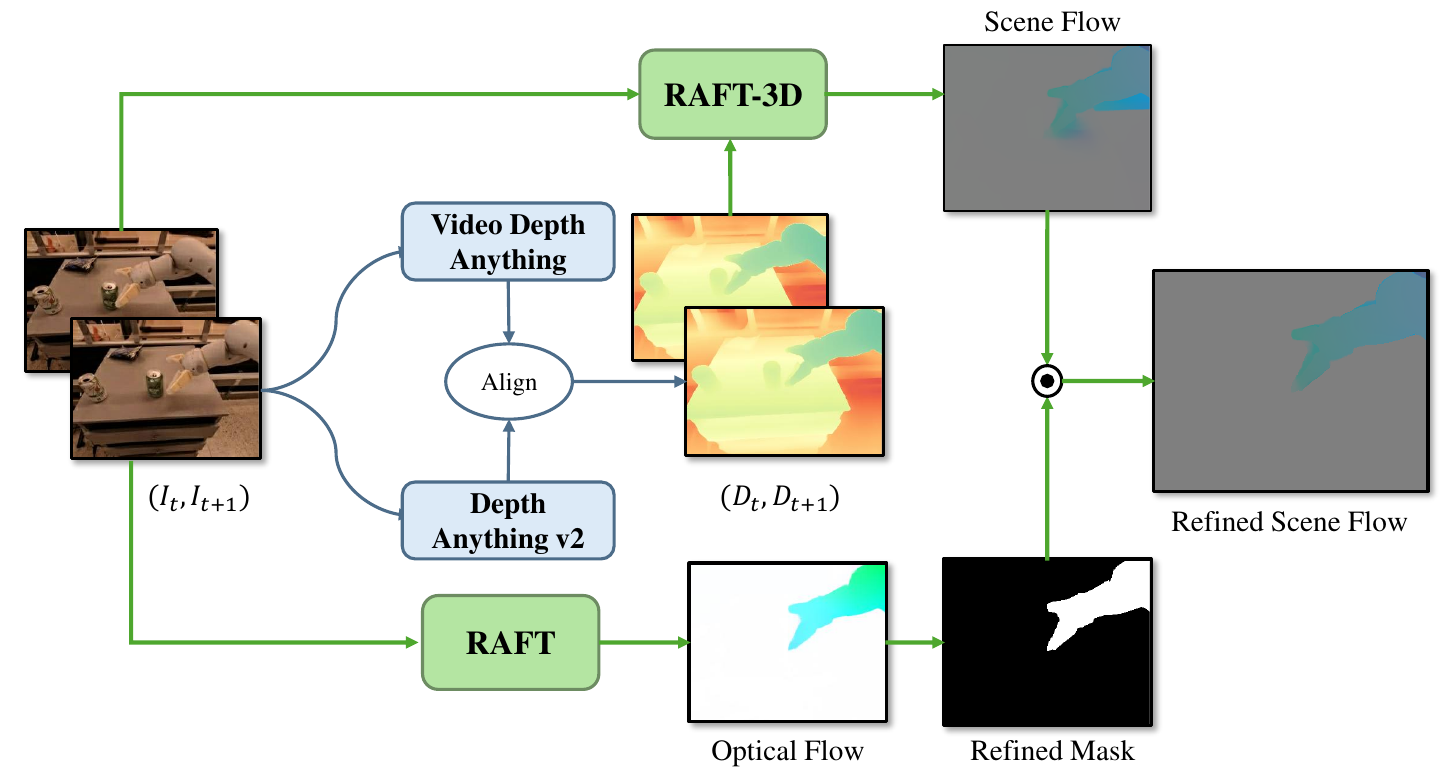}
\caption{\textbf{Scene flow obtainment pipeline on the real-world dataset.} The consistent metric depth is estimated by Video Depth Anything and Depth Anything v2, and the 3D scene flow is estimated by RAFT-3D and refined by RAFT.} 
\label{fig:data_collection}
\end{figure}

For RT-1 real world datasets, as there is no ground truth depth or pose matrix, we apply some depth estimation model: Depth Anything v2~\cite{depth_anything_v2} and Video Depth Anything~\cite{video_depth_anything} to estimate the metric depth, and utilize some flow estimation model: RAFT~\cite{flow_raft} and RAFT-3D~\cite{flow_raft3d} to estimate the 3D scene flow. The pipeline is illustrated in Fig.~\ref{fig:data_collection}. Specifically, we first obtain the relative depths using Video Depth Anything, which can produce spatially and temporally coherent depths across frames. To get metric depths, we utilize Depth Anything V2 to estimate the metric depths and align the two depth maps by computing scale and shift parameters. Next, we input the aligned metric depth along with the RGB image into RAFT-3D to obtain frame-paired 3D scene flows and use the estimation result of RAFT to refine the scene flow by producing a flow mask.

\begin{figure*}[!t]
\centering
\includegraphics[width=1.0\linewidth]{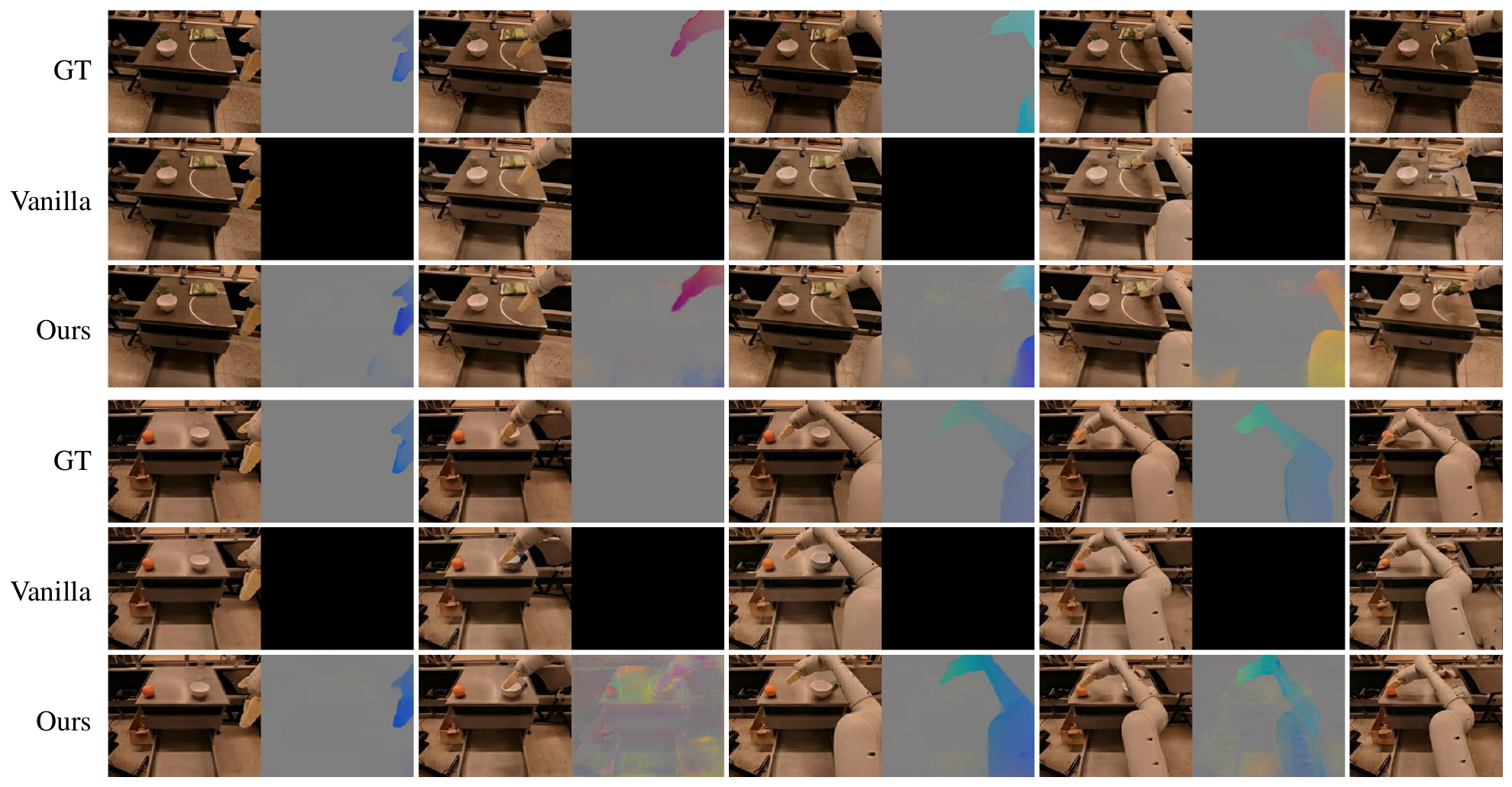}
\caption{\textbf{Qualitative results on the RT-1 real world dataset.} The depth is estimated by Depth Anything v2, and the flow in ``GT'' are estimated by RAFT-3D, making it a harder scenario.} 
\label{fig:qualitative_real_world}
\end{figure*}

\section{Extended Experimental Results}
\label{sec:extended_experiments}

% In this section, we provide more experimental results to evaluate \method, including real-world dataset experiments and comparison with state-of-the-art RGB-based world models.

% \subsection{Video Prediction on Real-world Dataset} 

In this section, we conduct real-world experiments to evaluate the feasibility of our pipeline. We conduct video prediction experiments on the real-world RT-1 robot manipulation dataset. The RT-1 real-world dataset contains more tasks, lighting conditions, and camera positions, making it a much harder task than on the simulation data. As there is no ground truth depth or scene flow, we leverage the estimation results as the training target. We compare \method with the single-stage \textit{Vanilla} world model.

Table~\ref{tab:video_prediction_real} shows the performance of video prediction. \method still performs better than \textit{Vanilla}, while the discrepancy becomes lower than simulation data. From the qualitative examples (see Fig.~\ref{fig:qualitative_real_world}), we could see that the ground truth scene flows estimated by RAFT-3D are not accurate, which would introduce noise into the training target. Our \method is affected by inaccurate supervision and produces inaccurate flow predictions. However, our performance still outperforms \textit{Vanilla}, as \textit{Vanilla} cannot even keep the consistency of the background during generation.

\section{Limitations and Future Works}

While \method has made progress, there are some limitations that could be improved by future works. First, our training data mainly come from simulation, as most of the robotic data are monocular RGB videos. Next, we did not consider any context or history in \method, though adding more context does not conflict with our methodology. We provide a simple version of \method that only relies on current observations and actions, just aiming to demonstrate the effectiveness of explicit dynamics modeling. Finally, the inference of the diffusion model is relatively slow due to the progressive denoising framework, while diffusion models could achieve outstanding visual performance. Future works could explore a better tradeoff between inference time and prediction performance.

% \subsection{Comparison with RGB-based World Models} 

% State-of-the-art world models based on RGB input and robot actions are often pretrained on large-scale robot datasets, \eg, Open-X Embodiment~\cite{open_x_embodiment}, or trained for a fairly long time, \eg, $\sim$2000 GPU hours. It is unfair for us to compare with them on prediction quality metrics, and this comparison cannot figure out the effectiveness of the flow prediction. Consequently, we only provide results with baseline RGB-D world models in the main paper, but we also show some comparisons with RGB-based world models here to clarify the performance. To compare \method with RGB-based world models, we choose two state-of-the-art world model, IRASim~\cite{action_irasim} and iVideoGPT~\cite{action_ivideogpt}:

% \begin{itemize}
%     \item \textit{IRASim} is a DiT-based~\cite{dit} latent video diffusion model, which uses frame-level adaptation to impose the action condition. IRASim predicts 15 future frames in parallel, with one initial frame and a trajectory of 15 actions.
%     \item \textit{iVideoGPT} is an autoregressive world model which tokenizes the input frames by VQGAN~\cite{vqgan} and pre-train the world model on large-scale datasets~\cite{open_x_embodiment, ssv2}. iVideoGPT achieves interactive prediction by predicting one frame at once conditioned on context frames and actions.
% \end{itemize}

% We use the official implementation and checkpoints of IRASim and iVideoGPT, finetune them on RT-1 SimplerEnv and Language Table dataset, and compare them with our \method.

\end{document}